\pdfoutput=1
\documentclass[letterpaper]{article}
\usepackage[preprint]{aaai2027}
\usepackage[hyphens]{url}
\usepackage{graphicx}
\newcommand{\doi}[1]{\url{https://doi.org/#1}}
\usepackage{natbib}
\usepackage{caption}
\usepackage{booktabs}
\usepackage{bussproofs}
\usepackage{makecell}
\usepackage{amsmath}
\usepackage{amssymb}
\usepackage{amsthm}
\usepackage{algorithm}
\usepackage{algorithmic}

\DeclareMathOperator*{\argmax}{\arg\!\max}

\newcommand{\leancop}{\textsf{leanCoP}}
\newcommand{\pttp}{\textsf{PTTP}}
\newcommand{\leantap}{\textsf{leanTaP}}
\newcommand{\meancop}{\textsf{meanCoP}}
\newcommand{\montecop}{\textsf{monteCoP}}
\newcommand{\leancopc}{\ifmmode\text{\textsf{leanCoP}}_{\mathrm{C}}\else\textsf{leanCoP}\ensuremath{_{\mathrm{C}}}\fi}
\newcommand{\meancopc}{\ifmmode\text{\textsf{meanCoP}}_{\mathrm{C}}\else\textsf{meanCoP}\ensuremath{_{\mathrm{C}}}\fi}
\newcommand{\rlcopc}{\ifmmode\text{\textsf{rlCoP}}_{\mathrm{C}}\else\textsf{rlCoP}\ensuremath{_{\mathrm{C}}}\fi}

\newcommand{\rlcop}{\mbox{\sf rlCoP}}
\newcommand{\plcop}{\mbox{\sf plCoP}}

\newcommand{\ileancop}{\mbox{\sf ileanCoP}}

\usepackage{tikz}%
\newtheorem{example}{Example}

\newcommand{\codelinks}{%
\begin{links}
  \link{Code}{https://github.com/fredrrom/connections}
\end{links}}

\newcommand{\acknowledgementsSection}{%
\section*{Acknowledgements}
This work was performed using resources provided by the Cambridge Service
for Data Driven Discovery (CSD3) operated by the University of Cambridge
Research Computing Service (\url{www.csd3.cam.ac.uk}), provided by Dell EMC
and Intel using Tier-2 funding from the Engineering and Physical Sciences
Research Council (capital grant EP/T022159/1), and DiRAC funding from the
Science and Technology Facilities Council (\url{www.dirac.ac.uk}).}

\begin{document}
\author{Fredrik R{\o}mming, Mantas Bak\v{s}ys, Martin S. Fixman, Sean B. Holden}
\affiliations{University of Cambridge\\
fr409@cam.ac.uk}
\title{Imitation Learning for Connection-Tableau Construction}

\maketitle              %
\begin{abstract}
An automated theorem prover builds a proof step by step, choosing at each
point what to add and what to remove. We cast this construction as a policy
acting in a transition system induced by a formal calculus, which fixes which
steps are sound: for clausal connection tableaux, \leancop{}-style search and
\plcop{}/\rlcop{}-style planning then become stateful policies over one
interface, and policy-learning methods apply directly. We equip such policies with a graph neural
network that scores proof edits from structure that transfers across problems,
train it by imitation learning from found proofs, and measure how performance
holds as we remove search scaffolding, from full symbolic backtracking to a
policy the network drives alone. Within a fixed step budget on M2k, MPTP2078-bushy, and
TPTP~v9.2.1, learned policies solve up to 46\% more problems than \leancop{}, and
reach proofs in an order of magnitude fewer steps.
\end{abstract}
\codelinks

\section{Introduction}
\label{sec:intro}

Proof calculi determine which proof steps are sound; proof procedures
determine which admissible steps are tried, retained, or undone. We study
whether these procedural choices can be learned across problems. We formalize
the distinction by letting the calculus induce a transition system and the
procedure act as a policy within it. We instantiate the framework for clausal
connection tableaux \cite{letz_model_2001}, whose non-confluent construction
makes control central, and with it the need to undo earlier choices.

This framing makes the vast policy-learning literature readily applicable.
Each problem induces its own transition system, so we train one policy by
imitation across problems and evaluate it zero-shot on unseen ones. Found proofs are the expert signal: replaying a
proof labels each intermediate proof object with its next edit, with no trace
of the surrounding failed search. We iteratively add newly found proofs to the
training set, and measure coverage and retention of known proofs as search
scaffolding is removed, within a fixed budget of \emph{steps}: transitions, an
implementation-independent measure of effort.

Connection tableaux, like resolution \cite{robinson_machine_1965}, reason over
clauses and complementary literals, but their proof object is one compact
tableau rather than a growing set of derived clauses. That compactness costs
proof confluence \cite{baumgartner_confluent_2000,farber_curiously_2023}: a
tableau grown by admissible rule applications alone can become impossible to
close, even when different choices would have led to a proof. Exploring the
space of proof objects therefore amounts to \emph{modifying} a tableau.

Existing connection provers can be read as different policies for organizing
these construction choices. The classical examples are built on or inspired
by ``Prolog technology'': \pttp{} \cite{stickel_prolog_1988},
\textsf{SETHEO} \cite{letz_setheo_1992}, \textsf{METEOR}
\cite{astrachan_meteors_1991}, \leantap{} \cite{beckert_leantap_1995}, and
\leancop{} \cite{otten_leancop_2003,otten_leancop_2008} construct tableaux in
the manner of model elimination \cite{loveland_mechanical_1968}, inheriting
Prolog's depth-first, program-order treatment of inference choice points,
with iterative deepening for completeness and proof enumeration.

This search is memory efficient (its whole state is one stack) but committed
to fixed choices whatever the problem: depth-first, in program order,
backtracking chronologically. \leancop{} turns these choices into fixed strategies such as
cut and clause reordering \cite{otten_restricting_2010}, and the learned
variants \textsf{MaLeCoP} and \textsf{FEMaLeCoP} reorder the alternatives
within them \cite{urban_malecop_2011,kaliszyk_femalecop_2015}. Later systems
abandon Prolog's stack to explore control more freely:
\meancop{} makes backtracking a design parameter
\cite{farber_curiously_2023}, and \montecop{} \cite{farber_machine_2021},
\rlcop{} \cite{kaliszyk_reinforcement_2018}, and \plcop{}
\cite{zombori_prolog_2020} replace it with a search tree guided by
Monte-Carlo statistics and learned priors. Graph neural networks have
been trained to guide connection provers directly
\cite{rawson_neurally-guided_2019,olsak_property_2020,rawson_lazycop_2021}.
In each case the learning is built into one prover's search procedure. Beyond
connection calculi, \textsf{TRAIL} separates a learning agent from a
saturation reasoner \cite{crouse_deep_2021}, where proof confluence leaves
nothing to undo: control is clause selection alone. We
zoom out and consider search a part of learnable policy: the lineage above
becomes a family of stateful policies, whose stacks, depth bounds, cuts, and
planning statistics are policy memory, not state.

\section{Preliminaries}
\label{sec:prelim}

\subsection{Transition Systems, Policies, and Demonstrations}
\label{subsubsec:rl}

Consider a deterministic transition system with states \(S\), actions \(A\),
and partial transition function \(T:S\times A\rightharpoonup S\). We write
\[
    \mathcal{A}(s)=\{a\in A\mid T(s,a)\text{ is defined}\}
\]
for the actions valid at state \(s\). A trajectory is a finite sequence of transitions
\[
    \tau = (s_0,a_0,s_1,\ldots,a_{n-1},s_n)
\]
such that \(a_t \in \mathcal{A}(s_t)\) and \(s_{t+1}=T(s_t,a_t)\). If the transition
system designates a goal set \(S^\checkmark\subseteq S\), then a trajectory is
successful when \(s_n\in S^\checkmark\).

Given a state \(s\), a policy chooses among the enabled actions in
\(\mathcal{A}(s)\). A deterministic memoryless (Markovian) policy is a map
\[
    \pi : S \to A,\qquad \pi(s) \in \mathcal{A}(s).
\]

Starting from an initial memory \(\mu_0\in\mathcal{M}_\pi\), a stateful policy
is specified by an action map and a memory-update map,
\[
    \pi_{\mathrm{mem}} : S \times \mathcal{M}_\pi \to A,
    \qquad
    U_\pi : \mathcal{M}_\pi \times S \times A
        \to \mathcal{M}_\pi,
\]
used as
\[
    \begin{aligned}
    a_t &= \pi_{\mathrm{mem}}(s_t,\mu_t) \in \mathcal{A}(s_t),\\
    \mu_{t+1} &= U_\pi(\mu_t,s_t,a_t).
    \end{aligned}
\]
Here \(\mu_t\) summarizes the trajectory seen by the policy so far. A fully
history-dependent policy is the special case where \(\mu_t\) stores the entire
trajectory \((s_0,a_0,\ldots,s_t)\).

Equivalently, since the transition system is deterministic once \(a_t\) is
chosen, the choice and memory update may be bundled as
\(\tilde{\pi}(s_t,\mu_t)=(a_t,\mu_{t+1})\). This is a memoryless policy over the
augmented state \((s_t,\mu_t)\); keeping \(\mu_t\) outside
\(S\) separates system states from policy bookkeeping.

Each policy so far is deterministic. A \emph{stochastic} policy instead draws
\(a_t\) from a distribution \(\pi(\cdot \mid s_t)\) over the enabled actions,
or \(\pi_{\mathrm{mem}}(\cdot \mid s_t,\mu_t)\) with memory.

Policy learning uses trajectories generated by some (potentially suboptimal) policy. We write
\(\beta\) for the \emph{behavior policy} that generated a trajectory and
\(\pi\) for the \emph{target policy} being learned or evaluated. If
\(\beta=\pi\), the data are on-policy; if \(\beta\ne\pi\), they are
off-policy. For stateful target policies, replaying a trajectory under
\(U_\pi\) gives the memory values \(\mu^\pi_t\) used as policy inputs during
training.

Learning methods differ in how they use these trajectories. Imitation learning
treats trajectory actions as demonstrated labels \cite{osa_algorithmic_2018}.
Behavioral cloning extracts samples \((x_t,a_t)\), where \(x_t=s_t\) for a
memoryless policy and \(x_t=(s_t,\mu^\pi_t)\) for a stateful policy, and fits
\(\pi\) by maximum likelihood, raising the probability it assigns each
demonstrated action \(a_t\) at input \(x_t\).
This supervised use of trajectories is often a sample-efficient starting point
because it supplies local action labels rather than relying on delayed reward
credit assignment \cite{pomerleau_efficient_1991}.
Reinforcement learning instead augments the transition system with a reward
function and updates the policy from reward-bearing experience
\cite{sutton_reinforcement_2018}.

\subsection{Connection Tableaux}
\label{subsec:cc}

We follow the tree-based presentation of clausal connection tableaux
\cite{letz_model_2001}. We assume basic first-order syntax. An atomic formula
\(A\) is built from predicate symbols and terms, terms are built from function
symbols and variables, and a literal is either \(A\) or \(\neg A\). We use
positive representation for clausification: input formulae are Herbrandized and transformed into equi-valid disjunctions of conjunctions of literals, so proofs are direct rather than refutations. The calculus is stated over the resulting matrix \(M\), a finite set of
clauses, each a finite set of literal occurrences.
For a literal \(L\), write \(\overline{L}\) for the literal obtained by changing
the polarity of \(L\). A \emph{connection} between literal occurrences \(L\) and
\(K\) is \(\sigma\)-complementary when
\(\sigma(L)=\sigma(\overline{K})\). The substitution \(\sigma\) is rigid: once a
unifier is chosen, it is applied to the whole partial tableau. A \emph{copy} of
a clause is obtained by consistently renaming its variables to fresh variables.

A connection tableau for \(M\) is a finite rooted tree. Non-root nodes are
labelled by literal occurrences from fresh copies of clauses in \(M\). A branch
is a path from the root to a leaf. A branch is closed under \(\sigma\) if it
contains a \(\sigma\)-complementary pair of literals. A tableau is closed if all
of its branches are closed. An unclosed leaf is an \emph{open goal}; the
\emph{active path} for that goal is the sequence of ancestor literals on the
branch.

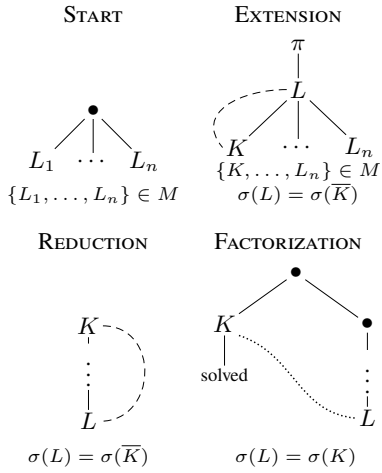
\begin{figure}[t]
\centering
\footnotesize
\setlength{\tabcolsep}{2pt}
\begin{tabular}{@{}cc@{}}
\textsc{Start} & \textsc{Extension} \\[1mm]
\begin{tikzpicture}[
    every node/.style={inner sep=1pt},
    edge/.style={line width=0.35pt},
    scale=0.9
]
    \node (root) at (0,0) {$\bullet$};
    \node (l1) at (-0.75,-0.75) {$L_1$};
    \node (dots) at (0,-0.75) {$\cdots$};
    \node (ln) at (0.75,-0.75) {$L_n$};
    \draw[edge] (root) -- (l1);
    \draw[edge] (root) -- (dots);
    \draw[edge] (root) -- (ln);
    \node[font=\scriptsize] at (0,-1.25) {$\{L_1,\ldots,L_n\}\in M$};
\end{tikzpicture}
&
\begin{tikzpicture}[
    every node/.style={inner sep=1pt},
    edge/.style={line width=0.35pt},
    connection/.style={densely dashed,line width=0.35pt},
    scale=0.9
]
    \node (anc) at (0,0.65) {\(\pi\)};
    \node (l) at (0,0) {$L$};
    \node (k) at (-0.9,-0.85) {$K$};
    \node (dots) at (0,-0.85) {$\cdots$};
    \node (ln) at (0.9,-0.85) {$L_n$};
    \draw[edge] (anc) -- (l);
    \draw[edge] (l) -- (k);
    \draw[edge] (l) -- (dots);
    \draw[edge] (l) -- (ln);
    \draw[connection] (l.west) to[out=180,in=130,looseness=1.35] (k.west);
    \node[font=\scriptsize,align=center] at (0,-1.38)
        {$\{K,\ldots,L_n\}\in M$\\[-1pt]
        $\sigma(L)=\sigma(\overline{K})$};
\end{tikzpicture}
\\[2mm]
\textsc{Reduction} & \textsc{Factorization} \\[1mm]
\begin{tikzpicture}[
    every node/.style={inner sep=1pt},
    edge/.style={line width=0.35pt},
    connection/.style={densely dashed,line width=0.35pt},
    scale=0.9
]
    \node (k) at (0,0.75) {$K$};
    \node (dots) at (0,0.15) {$\vdots$};
    \node (l) at (0,-0.65) {$L$};
    \draw[edge] (k) -- (dots);
    \draw[edge] (dots) -- (l);
    \draw[connection] (k.east) to[out=0,in=0,looseness=1.55] (l.east);
    \node[font=\scriptsize] at (0,-1.15) {$\sigma(L)=\sigma(\overline{K})$};
\end{tikzpicture}
&
\begin{tikzpicture}[
    every node/.style={inner sep=1pt},
    edge/.style={line width=0.35pt},
    factor/.style={densely dotted,line width=0.55pt},
    scale=0.9
]
    \node (n) at (0,0.65) {$\bullet$};
    \node (k) at (-1.05,-0.1) {$K$};
    \node (ns) at (1.05,-0.1) {$\bullet$};
    \node[font=\scriptsize] (solved) at (-1.05,-0.85) {solved};
    \node (dots) at (1.05,-0.82) {$\vdots$};
    \node (l) at (1.05,-1.48) {$L$};
    \draw[edge] (n) -- (k);
    \draw[edge] (n) -- (ns);
    \draw[edge] (k) -- (solved);
    \draw[edge] (ns) -- (dots);
    \draw[edge] (dots) -- (l);
    \draw[factor] (k.south east) to[out=-18,in=175,looseness=1.05] (l.west);
    \node[font=\scriptsize] at (0,-2.05) {$\sigma(L)=\sigma(K)$};
\end{tikzpicture}
\end{tabular}
\caption{Tree views of the start, extension, reduction, and factorization
operations. In extension and reduction, dashed lines mark closing connections
satisfying the displayed substitution equations. In factorization, the dotted
arc marks reuse of the solved subtableau rooted at the sibling \(K\).}
\label{fig:connection-tableau-rules}
\end{figure}

The calculus admits the following local operations, shown as tree fragments in
Figure~\ref{fig:connection-tableau-rules}:
\begin{itemize}
    \item \emph{Start:} choose a fresh copy \(C=\{L_1,\ldots,L_n\}\) of a
    clause in \(M\) and add a child of the root labelled \(L_i\) for each
    \(i\).
    \item \emph{Extension:} let the goal be labelled \(L\). Choose a
    fresh copy \(C=\{K,L_1,\ldots,L_n\}\) of a clause in \(M\). Extension is
    admissible when the current rigid substitution can be extended to
    \(\sigma\) with \(\sigma(L)=\sigma(\overline{K})\). One child labelled
    with each literal of \(C\) is added below the goal, and the child
    labelled \(K\) is closed by its connection with the goal.
    \item \emph{Reduction:} close an open goal labelled \(L\) if the current
    rigid substitution can be extended to \(\sigma\) with
    \(\sigma(L)=\sigma(\overline{K})\) for some literal \(K\) on its active
    path.
    \item \emph{Factorization:} close an open goal labelled \(L\) by reusing
    a solved node: a sibling of the goal or of one of its ancestors, labelled
    \(K\), such that the current rigid substitution can be extended to
    \(\sigma\) with \(\sigma(L)=\sigma(K)\). This is the pessimistic
    factorization of \citet{letz_model_2001}: since the source is already
    solved, the reuse cannot be circular, whereas the general rule admits
    unsolved sources and must carry a dependency ordering to keep two goals
    from justifying each other.
\end{itemize}

We also treat regularity as part of the construction system. A regularity
condition removes a transition step from consideration when it would create a
branch containing two equal same-polarity literals under the substitution that
step produces. Regularity is therefore an admissibility side condition on rule
applications.

The definitions above specify the proof objects, their closedness condition,
and the local rule applications; they do not prescribe an order in which open
goals or rule instances are tried.

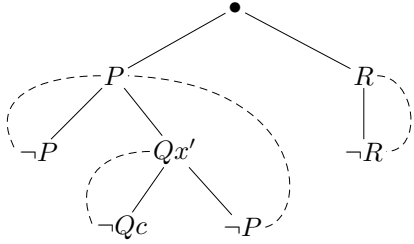
\begin{figure}[t]
\centering
\begin{tikzpicture}[
    every node/.style={inner sep=1pt},
    edge/.style={line width=0.35pt},
    connection/.style={densely dashed,line width=0.35pt}
]
    \node (root) at (0,0) {$\bullet$};
    \node (p) at (-1.6,-0.9) {$P$};
    \node (r) at (1.7,-0.9) {$R$};
    \node (np1) at (-2.6,-1.9) {$\neg P$};
    \node (qx) at (-0.8,-1.9) {$Qx'$};
    \node (nqc) at (-1.5,-2.9) {$\neg Qc$};
    \node (np2) at (0.1,-2.9) {$\neg P$};
    \node (nr) at (1.7,-1.9) {$\neg R$};

    \draw[edge] (root) -- (p);
    \draw[edge] (root) -- (r);
    \draw[edge] (p) -- (np1);
    \draw[edge] (p) -- (qx);
    \draw[edge] (qx) -- (nqc);
    \draw[edge] (qx) -- (np2);
    \draw[edge] (r) -- (nr);

    \draw[connection] (p.west) to[out=180,in=120,looseness=1.4] (np1.west);
    \draw[connection] (qx.west) to[out=180,in=115,looseness=1.35] (nqc.west);
    \draw[connection] (np2.east) .. controls (1.0,-2.75) and (0.95,-1.0) .. (p.east);
    \draw[connection] (r.east) to[out=0,in=0,looseness=1.3] (nr.east);
\end{tikzpicture}
\caption{A closed connection tableau for Example~\ref{exa:one}. 
Dashed curved lines mark closing connections between complementary
literals.}
\label{fig:connection-tableau-example}
\end{figure}

\begin{example}\label{exa:one}
Consider the formula
\[
F := ((P \vee \forall x~\neg(Qx \Rightarrow Qc)) \wedge R)
     \Rightarrow (P \wedge R).
\]
It has the matrix
\[
M =
\{\, \{P,R\}, \{\neg P,Qx\}, \{\neg P,\neg Qc\}, \{\neg R\} \,\}.
\]
One closed connection tableau starts with the clause \(\{P,R\}\), extends \(P\)
with \(\{\neg P,Qx'\}\), extends \(Qx'\) with \(\{\neg Qc,\neg P\}\) using
substitution \(\{x'\mapsto c\}\), reduces the remaining \(\neg P\) against the ancestor
\(P\), and closes \(R\) by extension with \(\{\neg R\}\), as shown in
Figure~\ref{fig:connection-tableau-example}.
\end{example}

\section{Connection-Tableau Construction as Control}
\label{sec:control}

Fix an input matrix \(M\). We model connection-tableau construction by the
proof-object transition system
\[
    \mathcal{P}(M) = (S, A, s_0, T, S^\checkmark).
\]
The states \(S\) are annotated partial connection tableaux for \(M\), carrying
the current rigid substitution and proof-object annotations such as rule
applications, parent links, and factorization sources. When we write ``the
tableau,'' we mean this transition-system state, with its substitution and
annotations left implicit. The initial state \(s_0\) is the
empty tableau, and \(S^\checkmark\subseteq S\) is the (potentially empty) set of closed tableaux.
Actions \(A\) are tableau edits. An edit either appends to the tableau by
applying an applicable calculus rule to an open goal, or prunes the tableau by
undoing the rule application at a node. Any rule-application node can be
pruned, and applications need not be undone in the reverse of the order in
which they were made. A prune removes the dependent rule applications and
resets the constraints they introduced; because the substitution is global,
dependence is not confined to the pruned subtree. The transition system therefore carries only logical
memory --- the tableau --- and no
control memory recording in what order the tableau was constructed.
The transition function \(T\) is partial because not every edit applies at
every state: it is defined exactly on the rule applications the calculus
admits and on undos of applications the tableau carries. We
distinguish an \emph{inference step}, a rule application recorded
in the proof object, from a \emph{transition step}, a single application of
\(T\): an append performs one inference step, while a prune removes one or
more. Step budgets and step counts in the evaluation refer to transition steps.

Action availability is read from the tableau itself: open goals, active
paths, the current substitution, and rule annotations determine which edits
are valid. Search-stack or frontier state, depth bounds, cuts, failed
alternatives, and planning statistics are control memory and belong to the
policy, which decides which valid edit to try next.

This separation gives a simple soundness guarantee. Any policy that acts only
through \(T\) produces a sequence of valid partial tableaux, and any terminal
state in \(S^\checkmark\) is a checkable closed tableau. Completeness is a
property of the policy or proof procedure: it must explore the transition system
fairly enough, subject to its bounds and pruning refinements, to find a closed
tableau whenever one is reachable. Forward connection-tableau construction is
not generally proof confluent: committing to one admissible rule application may
require later backtracking even if another rule application would have led to a
proof. The transition system, however, is reversible: every committed edit has
an inverse undo edit, so the state graph itself is confluent in the trivial
sense that diverging paths can return to their common predecessor. The stack,
frontier, or tree that decides which inverse to take, and which alternative to
try next, remains policy memory.

Consider the matrix from Example~\ref{exa:one}:
\[
M = \{ \{P, R\}, \{\neg P, Qx\}, \{\neg P, \neg Qc\}, \{\neg R\} \}.
\]
Figure~\ref{fig:state} shows a proof-object state \(s\) reachable by choosing
the start clause \(\{P,R\}\), then extending the open goal \(P\) with
\(\{\neg P,Qx'\}\), and then extending \(Qx'\) with
\(\{\neg Qc,\neg P\}\). The substitution \(\{x'\mapsto c\}\) is implicit in
the tableau state. At this point the tableau, the active paths, and the matrix
determine the enabled action set \(\mathcal{A}(s)\). Regularity is visible in
\(\mathcal{A}(s)\): the open goal \(\neg P\) unifies with the \(P\) of the
clause \(\{P,R\}\), but that extension would place a second \(P\) on its
branch and is therefore not enabled; the goal closes only by the reduction
against its ancestor \(P\).

\begin{figure}[t]
\centering
\begin{tikzpicture}[
    font=\small,
    every node/.style={inner sep=1pt},
    edge/.style={line width=0.35pt},
    connection/.style={densely dashed,line width=0.35pt},
    sectionrule/.style={black!35,line width=0.25pt},
    label/.style={anchor=west,font=\small,align=left},
    rowhead/.style={anchor=west,font=\small\bfseries\boldmath,align=left},
    action/.style={font=\small},
    lit/.style={font=\small,inner sep=0.5pt},
    note/.style={font=\small,align=center}
]
    \path (-3.55,-4.02) rectangle (3.55,2.42);

    \node[rowhead] at (-3.35,2.18) {\(M\):};
    \node[label] at (-2.65,2.18)
        {\(\{\{P,R\},\{\neg P,Qx\},\{\neg P,\neg Qc\},\{\neg R\}\}\)};
    \draw[sectionrule] (-3.35,1.78) -- (3.35,1.78);

    \node[rowhead] at (-3.35,1.42) {\(s\in S\)};

    \node (root) at (0,0.88) {$\bullet$};
    \node (p) at (-1.35,0.18) {$P$};
    \node (r) at (1.45,0.18) {$R$};
    \node (np1) at (-2.15,-0.58) {$\neg P$};
    \node (qx) at (-0.55,-0.58) {$Qx'$};
    \node (nqc) at (-1.15,-1.32) {$\neg Qc$};
    \node (np2) at (0.15,-1.32) {$\neg P$};

    \draw[edge] (root) -- (p);
    \draw[edge] (root) -- (r);
    \draw[edge] (p) -- (np1);
    \draw[edge] (p) -- (qx);
    \draw[edge] (qx) -- (nqc);
    \draw[edge] (qx) -- (np2);

    \draw[connection] (p.west) to[out=180,in=120,looseness=1.4] (np1.west);
    \draw[connection] (qx.west) to[out=180,in=115,looseness=1.35] (nqc.west);
    \draw[sectionrule] (-3.35,-1.72) -- (3.35,-1.72);

    \node[rowhead] at (-3.35,-2.08) {\(\mathcal{A}(s)\subseteq A\)};

    \node[action] at (-2.35,-2.58) {reduce};
    \node[lit] (redp) at (-2.57,-3.02) {$P$};
    \node[lit] (rednp) at (-2.02,-3.58) {$\neg P$};
    \draw[connection] (redp.east) to[out=0,in=70,looseness=1.2] (rednp.north);

    \node[action] at (-1.00,-2.58) {extend};
    \node[lit] (extr) at (-1.00,-3.02) {$R$};
    \node[lit] (extnr) at (-1.00,-3.58) {$\neg R$};
    \draw[edge] (extr) -- (extnr);
    \draw[connection] (extr.east) to[out=0,in=0,looseness=1.25] (extnr.east);

    \node[action] at (0.35,-2.58) {undo};
    \node[lit] (uq) at (0.35,-3.02) {$Qx'$};
    \node[lit] (uqc) at (-0.07,-3.58) {$\neg Qc$};
    \node[lit] (unp) at (0.80,-3.58) {$\neg P$};
    \draw[edge] (uq) -- (uqc);
    \draw[edge] (uq) -- (unp);

    \node[action] at (1.70,-2.58) {undo};
    \node[lit] (up) at (1.70,-3.02) {$P$};
    \node[lit] (upnp) at (1.30,-3.58) {$\neg P$};
    \node[lit] (upqx) at (2.11,-3.58) {$Qx'$};
    \draw[edge] (up) -- (upnp);
    \draw[edge] (up) -- (upqx);

    \node[action] at (3.05,-2.58) {undo};
    \node[lit] (uroot) at (3.05,-3.02) {$\bullet$};
    \node[lit] (usp) at (2.70,-3.58) {$P$};
    \node[lit] (usr) at (3.40,-3.58) {$R$};
    \draw[edge] (uroot) -- (usp);
    \draw[edge] (uroot) -- (usr);
\end{tikzpicture}
\caption{A single proof-object state and its enabled tableau-edit labels.
Dashed curved lines mark connections already made.}
\label{fig:state}
\end{figure}

\section{Learning from Proof-Trajectory Replay}
\label{sec:learning}

We consider the zero-shot multi-task problem of using proofs found in
\(\mathcal{P}(M)\) to find proofs in \(\mathcal{P}(M')\), and take the
sample-efficient imitation route, taking found proofs to generate
demonstrations. Given a successful trajectory \(\tau\) with final state
\(s_n \in S^\checkmark\) found by a behavior policy \(\beta\), we read off the
rule applications that construct the closed tableau in \(s_n\) and replay them from the
empty tableau relative to a target policy \(\pi\). If \(\pi\) has memory, replay
also runs its memory update. We define the resulting \emph{proof trajectory for \(\pi\)}
\[
    \tau^\pi =
    ((s_0,\mu_0^\pi),a_0^\star,(s_1,\mu_1^\pi),
    \ldots,a_{n-1}^\star,(s_n,\mu_n^\pi)),
\]
where \(s_{t+1}=T(s_t,a_t^\star)\), and
\[
    \mu_{t+1}^\pi = U_\pi(\mu_t^\pi,s_t,a_t^\star),
    \qquad
    a_t^\star\in\mathcal{A}(s_t,\mu_t^\pi).
\]
If at any step the recovered action is not in
\(\mathcal{A}(s_t,\mu_t^\pi)\), replay fails for that target policy: for
example, a policy whose memory exposes only actions within a growing depth
bound cannot replay a proof that closes a deep branch first. The closed
tableau may still be usable for another policy whose memory admits a different
construction order.
For a memoryless target policy, \(\mu_t^\pi\) is empty and
\(\mathcal{A}(s_t,\mu_t^\pi)=\mathcal{A}(s_t)\).

The behavior-policy run \(\tau\) that found the closed tableau may contain
failed branches, undo actions, and other search effort: as a demonstrator,
\(\beta\) is far from optimal. Replay discards this surrounding search and keeps
only the proof-producing actions, so the trajectory we imitate is not
\(\beta\)'s but that of an optimal expert \(\beta^\star\), the refinement of
\(\beta\) to the successful subset of its run, which reaches the closed tableau
with no wasted step. Unlike behavioral cloning from a fixed demonstrator, our
demonstrations therefore follow an optimal policy, and they are recovered
automatically: replay labels each \((s_t,\mu_t^\pi)\) with its next construction
action \(a_t^\star\), for closed tableaux alone.

Let \(\mathcal{D}\) be a dataset of replayed proof trajectories. \emph{Proof
cloning} (PC) trains on \(\mathcal{D}\) by supervised action prediction, as in
behavioral cloning \cite{pomerleau_efficient_1991}, decomposing each
trajectory into per-decision terms:
\[
    \mathcal{L}(\theta;\mathcal{D})
    =
    - \sum_{\tau\in\mathcal{D}}
      \sum_{t<n_\tau}
      \log \pi_\theta(a_t^\star\mid s_t,\mu_t^\pi).
\]
This \(\mathcal{L}(\theta;\mathcal{D})\) is the negative log-likelihood of the
demonstrated actions under \(\pi_\theta\), a cross-entropy loss.

Proof cloning suffers the distribution shift that limits behavioral cloning:
trained on the states of successful proofs alone, the policy has no guidance
once its own choices carry it elsewhere, and errors compound. \emph{Proof
aggregation} (PA) applies the standard DAgger remedy \cite{ross_reduction_2011},
retraining each round on the proofs the current policy itself finds
(Algorithm~\ref{alg:proof_aggregation}), so the training distribution follows
the states the policy visits. Our expert is weaker than DAgger's queryable
oracle, though: a proof labels only its own successful path, so unsolved
problems yield no labels and states not present in any proof path stay
unlabeled,
bounding the supervision the policy can receive. For the policy, this bound
may cost little: \plcop{}, training on Monte-Carlo search statistics from
every proof attempt, found policy data worth extracting only from searches
that found proofs \cite{zombori_prolog_2020}.

\begin{algorithm}[t]
\small
\caption{Proof Aggregation (PA)}
\label{alg:proof_aggregation}
\begin{algorithmic}
\STATE \textbf{Input:} matrices \(\mathcal{M}\), seed \(\pi_0\), target
    \(\pi\), rounds \(N\)
\STATE \(\mathcal{T} \gets \emptyset\) \quad\COMMENT{closed tableaux found so far}
\FOR{\(i \gets 1\) \textbf{to} \(N\)}
    \STATE \(\beta \gets \pi_{i-1}\) \quad\COMMENT{behavior policy of round \(i\)}
    \STATE Run \(\beta\) on \(\mathcal{M}\), adding closed tableaux to \(\mathcal{T}\)
    \STATE \(\mathcal{D} \gets \bigcup_{t\in\mathcal{T}} \mathrm{Replay}(t,\pi)\)
        \quad\COMMENT{proof trajectories for \(\pi\)}
    \STATE \(\pi_i \gets\) train \(\pi\) on \(\mathcal{D}\) by proof cloning
\ENDFOR
\STATE \textbf{Return:} final policy \(\pi_N\)
\end{algorithmic}
\end{algorithm}

\section{Implementation}
\label{sec:implementation}

\subsection{Prover Infrastructure}
\label{subsec:infrastructure}

Our implementation of the transition system and prover primitives builds on
and heavily extends earlier work by \citet{romming_connections_2023}.
A run takes a problem specification and a
schedule. The problem specification names the input file, a TPTP FOF/CNF
problem \citep{sutcliffe_tptp_2017} or a QMLTP QMF problem
\citep{raths_qmltp_2012}, together with the target logic and domain
semantics: classical or intuitionistic first-order logic, or first-order
modal logic over constant, cumulative, or varying domains; the evaluation
below is classical first-order throughout. The schedule
assigns step or time budgets to strategies and runs them in order until an
SZS \texttt{Success} or \texttt{NoSuccess} status is reported.

A strategy consists of a policy type together with matrix and policy options. The matrix options are the clausification
settings: they govern how a first-order (FOF) or modal (QMF) input formula becomes a
set of clauses, and have no effect on an input already in clausal form (CNF). We
support definitional \citep{plaisted_structure-preserving_1986} and naive
clause-form translation, corresponding to leanCoP's
\texttt{def} and \texttt{nodef} options respectively.
The matrix options thus fix the matrix, and with it the transition system. The
policy options determine how the policy is built and behaves. A strategy thus
pairs a policy type \(\Pi\), the policy that acts in the transition system,
with matrix options \(\gamma\) and policy options
\(\omega\); we call \((\gamma,\omega)\) an \emph{option profile}.

For our experiments, and to reproduce the leanCoP-family provers stepwise, our
matrix implementation retains the input order of clauses and their literals,
and we implement a hierarchy of memory update functions, \(U_{\mathrm{dfs}}\)
and \(U_{\mathrm{id}}\). \(U_{\mathrm{dfs}}\) maintains a stack of frames, each
holding the untried rule applications of the leftmost open goal and the undo edit that
abandons it; its top frame is the ordered action list
\(\mathcal{A}(s,\mu)\subseteq\mathcal{A}(s)\). \(U_{\mathrm{id}}\) restricts this
further by an iterative-deepening bound. The policy \(\pi_{\text{\leancop}}\)
runs on \(U_{\mathrm{id}}\) and takes the first action in \(\mathcal{A}(s,\mu)\), which, with the
input order retained, follows leanCoP's program order. leanCoP's search
arguments become policy options: \texttt{conj} restricts which
start actions are pushed to the stack, \texttt{cut} and \texttt{scut} discard
alternatives or frames from the stack, the latter for start clauses only, and
\texttt{comp(I)} lifts both above the depth bound. Factorization under
depth-first construction has its solved siblings exactly the nodes above and to
the left, and leanCoP restricts it further to equality: the reused literal must
already equal the goal under the current substitution, with no extension by
unification. That restricted form is leanCoP's \texttt{lemma}. We expose the
mode as a policy option and run every prover with factorization set to equality
rather than unification. We have
tested the implementation for stepwise equivalence to \leancop{}~2.1,
\ileancop{}~1.2, and \textsf{MleanCoP}~1.3 under the corresponding logic and
domain settings \citep{otten_leancop_2008,otten_mleancop_2014}.

Several caches keep transitions cheap. Terms, atoms, and literals
are immutable objects with precomputed hashes, so structural equality and
dictionary lookups do not re-traverse terms. The matrix is built once per
problem and matrix option set and shared by all strategies of a
schedule, together with its lazily refined \emph{connection graph}
\citep{bibel_deduction_1993}: a map from each literal to the literals it can
connect to, computed on first lookup as those statically unifiable with it,
that is, unifiable under the empty substitution, and cached for the rest of
the schedule. Each tableau node likewise caches its extension rules and
indexes its path and lemma candidates by signed predicate symbol, so
enumerating \(\mathcal{A}(s)\) rescans neither the matrix nor the path.
Transitions apply to a single mutable state and are reverted by inverse
edits rather than by copying.

\subsection{Learned Policies}
\label{subsec:policies}

The learned policies in our evaluation replace the first-action head of
\(\pi_{\text{\leancop}}\) with a scoring model that assigns a score
\(h_\theta(s,a)\) to each \(a\in\mathcal{A}(s,\mu)\). 
We consider three learned policies, each with increasingly restrictive
\(\mathcal{A}(s,\mu)\): \(\pi_{\mathrm{markov}}\), with \(\mathcal{A}(s,\mu) = \mathcal{A}(s)\),
\(\pi_{\mathrm{dfs}}\) with \(U_\mathrm{dfs}\), and \(\pi_{\mathrm{id}}\) 
with \(U_{\mathrm{id}}\). A learned strategy uses one of these as its \(\Pi\) in
place of \(\pi_{\text{\leancop}}\).

The policy is the softmax of the scores \(h_\theta\) over \(\mathcal{A}(s,\mu)\),
\[
    \pi_\theta(a\mid s,\mu)
    =
    \frac{\exp h_\theta(s,a)}
         {\sum_{a'\in\mathcal{A}(s,\mu)} \exp h_\theta(s,a')},
\]
a distribution we fit with the cross-entropy loss in proof cloning. Because this
normalization set \(\mathcal{A}(s,\mu)\) differs by memory, the three policy
types define different prediction problems, and we train a separate model for
each. Our evaluation runs are deterministic, committing to the mode
\[
    \pi_\theta(s,\mu)
    \in
    \argmax_{a\in\mathcal{A}(s,\mu)}
    \pi_\theta(a\mid s,\mu).
\]

\subsection{Scorer}
\label{subsec:scorer}

We implement the scorer \(h_\theta\) as a graph neural network over a typed
graph \(G(s)=(V,E)\). Its nodes, listed in Table~\ref{tab:nodes}, are of two
kinds: \emph{syntax nodes} carry the shared term language, while
\emph{occurrence nodes} mark positions in the proof. Everything but the goal
nodes and the relations involving them is fixed for the search and forms the
\emph{matrix} subgraph, encoded once per problem.

\begin{table}[t]
\small\centering
\begin{tabular}{@{}lll@{}}
\toprule
Kind & Node & One per \\
\midrule
syntax     & \emph{term}     & atom, compound term, or constant \\
           & \emph{variable} & variable \\
           & \emph{symbol}   & predicate, function, or constant \\
\addlinespace
occurrence & \emph{clause}   & clause of the matrix \\
           & \emph{literal}  & literal of the matrix \\
           & \emph{goal}     & goal of the tableau \\
\midrule
Relation & From & To \\
\midrule
\emph{membership} & clause          & its literals \\
\emph{atom}       & literal         & its atom \\
\emph{head}       & term            & its symbol \\
\emph{argument}   & term            & its subterms and variables \\
\emph{connection} & literal         & statically unifiable complements \\
\emph{instance}   & goal            & the matrix literal it instantiates \\
\emph{parent}     & goal            & its parent \\
\emph{path}       & goal            & each of its ancestors \\
\bottomrule
\end{tabular}
\caption{Node types and relations of \(G(s)\).}
\label{tab:nodes}
\end{table}

Syntax nodes are hash-consed: identical ground structure collapses to one node,
so the occurrences \(P(a)\) and \(\lnot P(a)\) share one term node, while
structure containing variables is shared only within a clause. A
node's initial features are attributes of its type alone: a symbol carries only
its kind and arity, never an identifier, and a variable carries none, taking
its identity from its edges alone. The encoding is therefore invariant to
signature renaming, and a trained scorer does not depend on the symbol names of
its training corpus, as in other symbol-independent proof guidance
\cite{olsak_property_2020,jakubuv_enigma_2020}.

Messages follow the relational graph-convolution form
\citep{schlichtkrull_modeling_2018}. Write \(N_r(v)\) for the \(r\)-neighbors of
a node \(v\). Each relation \(r\in R\) is directed and carries its own transform
\(W_r\); to let information flow both ways, \(R\) includes the reverse of every
edge type. The argument relation additionally adds a shared embedding
\(\rho_{uv}\) of the argument position to the neighbor before \(W_r\), with
\(\rho_{uv}=0\) on the other relations. A round first aggregates these messages
by a mean within each relation and a sum across relations,
\[
    m_v^{(\ell)} = \sum_{r\in R}\frac{1}{\lvert N_r(v)\rvert}
        \sum_{u\in N_r(v)} W_r\big(h_u^{(\ell)} + \rho_{uv}\big),
\]
then updates each node with a residual around a self-transform
\(W^{\mathrm{self}}_{\phi(v)}\) and a layer norm \(\mathrm{LN}_{\phi(v)}\), both
keyed to the node's type \(\phi(v)\),
\[
    h_v^{(\ell+1)} = \mathrm{LN}_{\phi(v)}\big(
        h_v^{(\ell)} + \tanh(W^{\mathrm{self}}_{\phi(v)}\, h_v^{(\ell)}
        + m_v^{(\ell)})\big).
\]
Initial embeddings \(h_v^{(0)}\) sum a learned embedding of each of the node's
structural features and apply a tanh. The transforms are shared across rounds; after \(L\) rounds we
read off the node embeddings \(e_v := h_v^{(L)}\).

Every candidate edit \(a\) then has an edit type \(k(a)\) and an ordered
endpoint pair whose embeddings feed a shared head, and that pair is an edge
already present: the source is always the acting goal, and the target is the
start clause for a start, the matrix literal for an extension, and the path
ancestor for a reduction or lemma, while a backtrack has no target and
takes a learned null embedding \(e_\varnothing\). With \(\kappa_{k(a)}\) a
learned embedding of the edit type and \(e_{\mathrm{src}},e_{\mathrm{tgt}}\) the
endpoint embeddings,
\[
    h_\theta(s,a) = f_\theta\big([\kappa_{k(a)}; e_{\mathrm{src}}; e_{\mathrm{tgt}}]\big),
\]
where \(f_\theta\) is a multilayer perceptron with two tanh hidden layers and a
scalar output.

The encoding stays cheap to update because message flow is directional.
The matrix subgraph is encoded on its own, and the goals are then encoded with
the matrix embeddings held fixed, so those embeddings do not depend on
the tableau. Computed once at the first decision, the matrix embeddings are
exact for every later decision and reused; only the goals are re-encoded as the
tableau grows. The same sharing applies across a training batch, where all
trajectories from one problem share a single copy of its matrix subgraph.

\section{Evaluation}
\label{sec:experiments}

We evaluate the effect of learning by holding the calculus, transition
interface, and option profile fixed and varying only the policy, so any
difference isolates the learned choice rather than a difference between
provers. We therefore do not compare against other connection provers or
saturation systems, where the learned component would be confounded with
differences in calculus, strategy, and implementation language; the aim is not
yet a competition-winning prover. We test on the
Kaliszyk--Urban Mizar line, M2k and MPTP2078-bushy (MPTP)
\citep{kaliszyk_reinforcement_2018,kaliszyk_femalecop_2015}, and on the FOF
problems of TPTP~v9.2.1 \citep{sutcliffe_tptp_2017}. A problem counts as solved when a run
reaches a closed tableau: SZS \texttt{Theorem} when the problem has a
conjecture and \texttt{Unsatisfiable} when it does not. Exhausting the search
space instead yields \texttt{CounterSatisfiable} or \texttt{Satisfiable}, which
give no replay signal and are not counted. Effort is reported as ``Steps'', the average
number of transitions over successful runs.

We evaluate five leanCoP option profiles from the restricted-backtracking
ablation grid \cite{otten_restricting_2010}, chosen so that every option axis is
exercised at least once: the bare \texttt{default}, conjecture start clauses
(\texttt{conj}), backtracking restriction alone (\texttt{cut}) and combined with
iterative-deepening completeness bounds (\texttt{cut,comp(7)}), and definitional
clausification with start-clause restriction (\texttt{def,scut}). At each
profile, \(\pi_{\text{\leancop}}\) is the baseline, and the learned
\(\pi_{\mathrm{markov}}\), \(\pi_{\mathrm{dfs}}\), and \(\pi_{\mathrm{id}}\)
replace its first-action choice with the scorer over their respective memories.
Initial trajectories are extracted from \(\pi_{\text{\leancop}}\)'s closed
tableaux, and learned policies are evaluated after five proof-aggregation
iterations.

The training set follows from proof gathering: a problem
enters it only once a closed tableau for it has been replayed into
a trajectory and added to \(\mathcal{D}\). Evaluation is coverage on the fixed
problem set, and there is no leakage: every coverage gain is a first solve,
made before the problem had contributed any training data. Whether a policy still solves the problems it trained on
is measured separately, as retention. At each iteration the scorer is
retrained from scratch on the deduplicated \((s,a)\) samples of all
trajectories gathered so far. A star search on the M2k \texttt{cut,comp(7)} profile, one axis at a time and
ranked by problems solved after two aggregation iterations, sets the
hyperparameters over width \(\in\{48,64,96\}\), learning rate
\(\in\{3\times10^{-4},10^{-3},3\times10^{-3}\}\), and batch size
\(\in\{32,64,128\}\); the centre won or tied on every axis, giving width 64
for all embeddings and hidden layers, Adam at learning rate \(10^{-3}\), and
batch size 64, trained until five epochs pass with no improvement in training
loss. There is no held-out validation set. Under imitation on a fixed task set, fitting the demonstrations
exactly is the goal rather than overfitting, with generalization measured as
transfer to unsolved problems. Training is the only stochastic component of
the system, and since it ends at near-perfect accuracy the seed is immaterial;
we fix seed 0, and evaluation runs are deterministic. The depth \(L\) follows a
Weisfeiler--Leman analysis of an M2k training set: a message-passing network is
no more discriminating than \(L\) rounds of color refinement
\cite{xu_how_2019}, and three rounds separate every labeled action from the
competing actions of its edit type, so \(L=4\). Across the 225 trainings
reported here mean training accuracy is 0.975, and the datasets grow from a
few thousand \((s,a)\) samples to a median of 12{,}000--16{,}000.
All runs, baseline and learned alike, use a budget of 1{,}000 transition steps
and a 120-second wall-clock guard, on one Intel Xeon Platinum core under a
3\,GB memory limit; exhausting any of these counts as unsolved. The system
runs on Rocky Linux~8.10 with PyTorch~2.10 on Python~3.12.

Table~\ref{tab:generalization} reports solved counts across corpora and option
profiles. A learned policy beats the baseline in every row, so the gain is not
tied to a particular problem distribution or search configuration.
\(\pi_{\mathrm{dfs}}\) and \(\pi_{\mathrm{id}}\) are the strongest, taking
about half the rows each, while \(\pi_{\mathrm{markov}}\) is inconsistent and
falls below the baseline in several.
\begin{table}[t]
\centering
\setlength{\tabcolsep}{3.5pt}
\begin{tabular}{@{}llrrrr@{}}
\toprule
Corpus & \((\gamma,\omega)\) &
\(\pi_{\text{\leancop}}\) & \(\pi_{\mathrm{markov}}\) &
\(\pi_{\mathrm{dfs}}\) & \(\pi_{\mathrm{id}}\) \\
\midrule
M2k & \texttt{default} & 544 & 554 & \textbf{729} & 670 \\
& \texttt{conj} & 474 & 531 & \textbf{694} & 547 \\
& \texttt{cut} & 602 & 583 & 738 & \textbf{750} \\
& \texttt{cut,comp(7)} & 594 & 585 & \textbf{756} & 726 \\
& \texttt{def,scut} & 532 & 560 & \textbf{625} & 622 \\
\midrule
MPTP & \texttt{default} & 255 & 295 & \textbf{316} & 310 \\
& \texttt{conj} & 259 & 287 & \textbf{312} & 307 \\
& \texttt{cut} & 282 & 279 & 339 & \textbf{365} \\
& \texttt{cut,comp(7)} & 281 & 305 & \textbf{363} & 340 \\
& \texttt{def,scut} & 252 & 234 & 301 & \textbf{304} \\
\midrule
TPTP & \texttt{default} & 898 & 898 & 1012 & \textbf{1013} \\
& \texttt{conj} & 999 & 861 & 997 & \textbf{1126} \\
& \texttt{cut} & 981 & 922 & 1056 & \textbf{1097} \\
& \texttt{cut,comp(7)} & 980 & 873 & 1077 & \textbf{1096} \\
& \texttt{def,scut} & 904 & 899 & 988 & \textbf{1080} \\
\bottomrule
\end{tabular}
\caption{Solved-problem coverage by corpus and option profile
\((\gamma,\omega)\), with early stopping.}
\label{tab:generalization}
\end{table}

Fixing \((\gamma,\omega)\) to \texttt{cut,comp(7)}, the ``early stopping''
block of Table~\ref{tab:fixed-strategy} counts, against the
\(\pi_{\text{\leancop}}\) baseline, the proofs each policy newly finds and the
ones it no longer finds. The three order by action space.
\(\pi_{\mathrm{markov}}\) reaches proofs in the fewest steps, an order of
magnitude below the baseline, but forfeits the most; \(\pi_{\mathrm{dfs}}\)
finds the most proofs and the most new ones at a step count far closer to it
than to \(\pi_{\mathrm{id}}\);
\(\pi_{\mathrm{id}}\) retains almost every baseline proof but spends the most
steps, still fewer than the baseline. Under early stopping, widening the
action space trades retention for shorter searches.

Coverage at this profile peaks in the middle rather than at either extreme,
and the training signal explains why: proof trajectories
contain the successful path alone, so undo edits never appear as labels and a
policy that must decide when to undo is never trained for it.
\(\pi_{\mathrm{dfs}}\) keeps the backtracking the trajectories omit while
remaining free enough not to re-derive the same prefixes at every bound;
\(\pi_{\mathrm{markov}}\) inherits none of it. Training on search traces
rather than proofs alone would supply that signal.

Everything so far stops training after five epochs without improvement, which
leaves the scorer, the only learned component, short of fitting its data. We
therefore retrain every model on MPTP to 100\% training accuracy or a 200-epoch
cap, holding calculus, option profile, trajectories, and evaluation fixed;
mean training accuracy rises from 0.975 to 0.995. Only MPTP has both protocols
run end to end.

\begin{table}[t]
\centering
\begin{tabular}{@{}llrrrr@{}}
\toprule
Training & Policy & Solv. & New & Lost & Steps \\
\midrule
--- & \(\pi_{\text{\leancop}}\) & 281 & 0 & 0 & 188.4 \\
\midrule
early stopping & \(\pi_{\mathrm{markov}}\) & 305 & 51 & 27 & 11.9 \\
& \(\pi_{\mathrm{dfs}}\) & 363 & 94 & 12 & 21.7 \\
& \(\pi_{\mathrm{id}}\) & 340 & 60 & 1 & 158.7 \\
\midrule
to convergence & \(\pi_{\mathrm{markov}}\) & 339 & 59 & 1 & 11.3 \\
& \(\pi_{\mathrm{dfs}}\) & \textbf{396} & \textbf{118} & 3 & 26.6 \\
& \(\pi_{\mathrm{id}}\) & 345 & 65 & 1 & 155.0 \\
\bottomrule
\end{tabular}
\caption{New/lost tradeoff on MPTP at
\((\gamma,\omega)=\texttt{cut,comp(7)}\), under both training protocols.
``New'' and ``Lost'' are relative to \(\pi_{\text{\leancop}}\); ``Steps''
averages over successful runs.}
\label{tab:fixed-strategy}
\end{table}

The second block of Table~\ref{tab:fixed-strategy} shows what that achieves. Every
policy finds more new proofs and solves more overall, and the largest change is
in ``Lost'': the proofs \(\pi_{\mathrm{markov}}\) forfeits fall from 27 to 1
and \(\pi_{\mathrm{dfs}}\)'s from 12 to 3, while \(\pi_{\mathrm{markov}}\)
keeps its order-of-magnitude step advantage. Most of that retention penalty is
therefore not intrinsic to a wide action space but scorer error, and largely
disappears once the scorer is fitted. Step counts
barely move except for \(\pi_{\mathrm{dfs}}\), whose rise from 21.7 to 26.6
comes from the longer proofs it newly finds: of the 65 problems it gains,
all but one previously exhausted the step or time budget.

The ordering persists, and the size of the effect follows it: training to
convergence is worth \(+34\) solved problems to \(\pi_{\mathrm{markov}}\) and
\(+33\) to \(\pi_{\mathrm{dfs}}\), but only \(+5\) to \(\pi_{\mathrm{id}}\).
The gain from fitting the scorer measures how much of a policy's behaviour
rests on learning rather than imposed search.

\begin{figure}[t]
\centering
\includegraphics{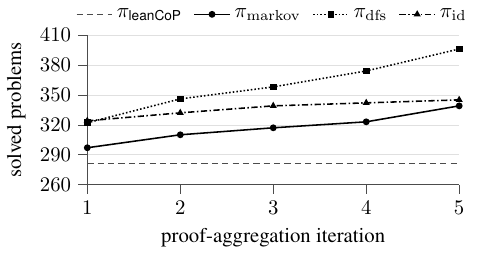}
\caption{Solved coverage over five aggregation iterations at
\((\gamma,\omega)=\texttt{cut,comp(7)}\) on MPTP, trained to convergence.}
\label{fig:aggregation-curve}
\end{figure}

Figure~\ref{fig:aggregation-curve} traces the converged policies through the
aggregation iterations. All three clear the baseline from the first iteration
and then level off, as the problems still unsolved lie outside the patterns
the gathered proofs cover or beyond the step budget.

\section{Conclusion}
\label{sec:conclusion}

We tackled the zero-shot multi-task problem of using proofs found on some
problems to improve prover performance on others. Casting connection-tableau
construction as policy learning over a transition system that admits only
calculus-valid edits leaves soundness with the system, while existing provers
employing search and planning can be seen as stateful policies interacting
with it. Replaying a proof labels each of its inference steps as a transition
step, and proof cloning with proof aggregation trains a graph-neural-network
policy on those labels. Within a fixed step budget across three benchmarks,
the gains differ by the policies' imposed search:
\(\pi_{\mathrm{id}}\) retains nearly every baseline
proof, \(\pi_{\mathrm{dfs}}\) finds the most new proofs and solves up to 46\%
more than the baseline, and \(\pi_{\mathrm{markov}}\) reaches proofs in an
order of magnitude fewer steps. Under early stopping, less imposed structure
buys shorter proofs at the cost of retention; fitting the scorer largely
recovers it.

\ifdefined\acknowledgementsSection\acknowledgementsSection\fi
\bibliography{references}

\begin{thebibliography}{34}
\providecommand{\natexlab}[1]{#1}

\bibitem[{Astrachan and Loveland(1991)}]{astrachan_meteors_1991}
Astrachan, O.~L.; and Loveland, D.~W. 1991.
\newblock {METEORs}: {High} {Performance} {Theorem} {Provers} {Using} {Model}
  {Elimination}.
\newblock In Boyer, R.~S., ed., \emph{Automated {Reasoning}: {Essays} in
  {Honor} of {Woody} {Bledsoe}}, volume~1 of \emph{Automated {Reasoning}
  {Series}}, 31--59. Dordrecht: Kluwer Academic Publishers.

\bibitem[{Baumgartner, Eisinger, and
  Furbach(2000)}]{baumgartner_confluent_2000}
Baumgartner, P.; Eisinger, N.; and Furbach, U. 2000.
\newblock A {Confluent} {Connection} {Calculus}.
\newblock In Hölldobler, S., ed., \emph{Intellectics and {Computational}
  {Logic}: {Papers} in {Honor} of {Wolfgang} {Bibel}}, volume~19 of
  \emph{Applied {Logic} {Series}}, 3--26. Dordrecht: Kluwer Academic
  Publishers.

\bibitem[{Beckert and Posegga(1995)}]{beckert_leantap_1995}
Beckert, B.; and Posegga, J. 1995.
\newblock {leanTAP}: {Lean} {Tableau}-{Based} {Deduction}.
\newblock \emph{Journal of Automated Reasoning}, 15(3): 339--358.

\bibitem[{Bibel(1993)}]{bibel_deduction_1993}
Bibel, W. 1993.
\newblock \emph{Deduction: automated logic}.
\newblock London: Academic Press.

\bibitem[{Crouse et~al.(2021)Crouse, Abdelaziz, Makni, Whitehead, Cornelio,
  Kapanipathi, Srinivas, Thost, Witbrock, and Fokoue}]{crouse_deep_2021}
Crouse, M.; Abdelaziz, I.; Makni, B.; Whitehead, S.; Cornelio, C.; Kapanipathi,
  P.; Srinivas, K.; Thost, V.; Witbrock, M.; and Fokoue, A. 2021.
\newblock A {Deep} {Reinforcement} {Learning} {Approach} to {First}-{Order}
  {Logic} {Theorem} {Proving}.
\newblock \emph{Proceedings of the AAAI Conference on Artificial Intelligence},
  35(7): 6279--6287.

\bibitem[{Färber(2023)}]{farber_curiously_2023}
Färber, M. 2023.
\newblock A {Curiously} {Effective} {Backtracking} {Strategy} for {Connection}
  {Tableaux}.
\newblock In Otten, J.; and Bibel, W., eds., \emph{Proceedings of the 1st
  {International} {Workshop} on {Automated} {Reasoning} with {Connection}
  {Calculi} ({AReCCa} 2023)}, volume 3613 of \emph{{CEUR} {Workshop}
  {Proceedings}}, 23--40. Prague, Czech Republic: CEUR.

\bibitem[{Färber, Kaliszyk, and Urban(2021)}]{farber_machine_2021}
Färber, M.; Kaliszyk, C.; and Urban, J. 2021.
\newblock Machine {Learning} {Guidance} for {Connection} {Tableaux}.
\newblock \emph{Journal of Automated Reasoning}, 65(2): 287--320.

\bibitem[{Jakubův et~al.(2020)Jakubův, Chvalovský, Olšák, Piotrowski,
  Suda, and Urban}]{jakubuv_enigma_2020}
Jakubův, J.; Chvalovský, K.; Olšák, M.; Piotrowski, B.; Suda, M.; and
  Urban, J. 2020.
\newblock {ENIGMA} {Anonymous}: {Symbol}-{Independent} {Inference} {Guiding}
  {Machine} ({System} {Description}).
\newblock In Peltier, N.; and Sofronie-Stokkermans, V., eds., \emph{Automated
  {Reasoning}}, volume 12167 of \emph{Lecture {Notes} in {Computer} {Science}},
  448--463. Cham: Springer International Publishing.

\bibitem[{Kaliszyk and Urban(2015)}]{kaliszyk_femalecop_2015}
Kaliszyk, C.; and Urban, J. 2015.
\newblock {FEMaLeCoP}: {Fairly} {Efficient} {Machine} {Learning} {Connection}
  {Prover}.
\newblock In Davis, M.; Fehnker, A.; McIver, A.; and Voronkov, A., eds.,
  \emph{Logic for {Programming}, {Artificial} {Intelligence}, and {Reasoning}},
  volume 9450 of \emph{Lecture {Notes} in {Computer} {Science}}, 88--96.
  Berlin, Heidelberg: Springer.

\bibitem[{Kaliszyk et~al.(2018)Kaliszyk, Urban, Michalewski, and
  Olšák}]{kaliszyk_reinforcement_2018}
Kaliszyk, C.; Urban, J.; Michalewski, H.; and Olšák, M. 2018.
\newblock Reinforcement {Learning} of {Theorem} {Proving}.
\newblock In \emph{Advances in {Neural} {Information} {Processing} {Systems}},
  volume~31, 8836--8847. Curran Associates, Inc.

\bibitem[{Letz et~al.(1992)Letz, Schumann, Bayerl, and
  Bibel}]{letz_setheo_1992}
Letz, R.; Schumann, J.; Bayerl, S.; and Bibel, W. 1992.
\newblock {SETHEO}: {A} high-performance theorem prover.
\newblock \emph{Journal of Automated Reasoning}, 8(2): 183--212.

\bibitem[{Letz and Stenz(2001)}]{letz_model_2001}
Letz, R.; and Stenz, G. 2001.
\newblock Model elimination and connection tableau procedures.
\newblock In \emph{Handbook of automated reasoning}, volume~II, 2015--2114.
  Elsevier and MIT Press.

\bibitem[{Loveland(1968)}]{loveland_mechanical_1968}
Loveland, D.~W. 1968.
\newblock Mechanical {Theorem}-{Proving} by {Model} {Elimination}.
\newblock \emph{Journal of the ACM}, 15(2): 236--251.

\bibitem[{Olšák, Kaliszyk, and Urban(2020)}]{olsak_property_2020}
Olšák, M.; Kaliszyk, C.; and Urban, J. 2020.
\newblock Property {Invariant} {Embedding} for {Automated} {Reasoning}.
\newblock In \emph{{ECAI} 2020}, volume 325 of \emph{Frontiers in {Artificial}
  {Intelligence} and {Applications}}, 1395--1402. IOS Press.

\bibitem[{Osa et~al.(2018)Osa, Pajarinen, Neumann, Bagnell, Abbeel, and
  Peters}]{osa_algorithmic_2018}
Osa, T.; Pajarinen, J.; Neumann, G.; Bagnell, J.~A.; Abbeel, P.; and Peters, J.
  2018.
\newblock An {Algorithmic} {Perspective} on {Imitation} {Learning}.
\newblock \emph{Foundations and Trends® in Robotics}, 7(1-2): 1--179.

\bibitem[{Otten(2008)}]{otten_leancop_2008}
Otten, J. 2008.
\newblock {leanCoP} 2.0 and {ileanCoP} 1.2: {High} {Performance} {Lean}
  {Theorem} {Proving} in {Classical} and {Intuitionistic} {Logic} ({System}
  {Descriptions}).
\newblock In Armando, A.; Baumgartner, P.; and Dowek, G., eds., \emph{Automated
  {Reasoning}}, volume 5195 of \emph{Lecture {Notes} in {Computer} {Science}},
  283--291. Berlin, Heidelberg: Springer.

\bibitem[{Otten(2010)}]{otten_restricting_2010}
Otten, J. 2010.
\newblock Restricting backtracking in connection calculi.
\newblock \emph{AI Communications}, 23(2-3): 159--182.

\bibitem[{Otten(2014)}]{otten_mleancop_2014}
Otten, J. 2014.
\newblock {MleanCoP}: {A} {Connection} {Prover} for {First}-{Order} {Modal}
  {Logic}.
\newblock In Demri, S.; Kapur, D.; and Weidenbach, C., eds., \emph{Automated
  {Reasoning}}, volume 8562 of \emph{Lecture {Notes} in {Computer} {Science}},
  269--276. Cham: Springer International Publishing.

\bibitem[{Otten and Bibel(2003)}]{otten_leancop_2003}
Otten, J.; and Bibel, W. 2003.
\newblock {leanCoP}: lean connection-based theorem proving.
\newblock \emph{Journal of Symbolic Computation}, 36(1--2): 139--161.

\bibitem[{Plaisted and Greenbaum(1986)}]{plaisted_structure-preserving_1986}
Plaisted, D.~A.; and Greenbaum, S. 1986.
\newblock A {Structure}-preserving {Clause} {Form} {Translation}.
\newblock \emph{Journal of Symbolic Computation}, 2(3): 293--304.

\bibitem[{Pomerleau(1991)}]{pomerleau_efficient_1991}
Pomerleau, D.~A. 1991.
\newblock Efficient {Training} of {Artificial} {Neural} {Networks} for
  {Autonomous} {Navigation}.
\newblock \emph{Neural Computation}, 3(1): 88--97.

\bibitem[{Raths and Otten(2012)}]{raths_qmltp_2012}
Raths, T.; and Otten, J. 2012.
\newblock The {QMLTP} {Problem} {Library} for {First}-{Order} {Modal} {Logics}.
\newblock In Gramlich, B.; Miller, D.; and Sattler, U., eds., \emph{Automated
  {Reasoning}}, volume 7364 of \emph{Lecture {Notes} in {Computer} {Science}},
  454--461. Berlin, Heidelberg: Springer.

\bibitem[{Rawson and Reger(2019)}]{rawson_neurally-guided_2019}
Rawson, M.; and Reger, G. 2019.
\newblock A {Neurally}-{Guided}, {Parallel} {Theorem} {Prover}.
\newblock In Herzig, A.; and Popescu, A., eds., \emph{Frontiers of {Combining}
  {Systems}}, volume 11715 of \emph{Lecture {Notes} in {Computer} {Science}},
  40--56. Cham: Springer International Publishing.

\bibitem[{Rawson and Reger(2021)}]{rawson_lazycop_2021}
Rawson, M.; and Reger, G. 2021.
\newblock {lazyCoP}: {Lazy} {Paramodulation} {Meets} {Neurally} {Guided}
  {Search}.
\newblock In Das, A.; and Negri, S., eds., \emph{Automated {Reasoning} with
  {Analytic} {Tableaux} and {Related} {Methods}}, volume 12842 of \emph{Lecture
  {Notes} in {Computer} {Science}}, 187--199. Cham: Springer International
  Publishing.

\bibitem[{Robinson(1965)}]{robinson_machine_1965}
Robinson, J.~A. 1965.
\newblock A {Machine}-{Oriented} {Logic} {Based} on the {Resolution}
  {Principle}.
\newblock \emph{Journal of the ACM}, 12(1): 23--41.

\bibitem[{Ross, Gordon, and Bagnell(2011)}]{ross_reduction_2011}
Ross, S.; Gordon, G.; and Bagnell, D. 2011.
\newblock A {Reduction} of {Imitation} {Learning} and {Structured} {Prediction}
  to {No}-{Regret} {Online} {Learning}.
\newblock In \emph{Proceedings of the {Fourteenth} {International} {Conference}
  on {Artificial} {Intelligence} and {Statistics}}, 627--635. JMLR Workshop and
  Conference Proceedings.

\bibitem[{Rømming, Otten, and Holden(2023)}]{romming_connections_2023}
Rømming, F.; Otten, J.; and Holden, S.~B. 2023.
\newblock Connections: {Markov} {Decision} {Processes} for {Classical},
  {Intuitionistic} and {Modal} {Connection} {Calculi}.
\newblock In Otten, J.; and Bibel, W., eds., \emph{Proceedings of the 1st
  {International} {Workshop} on {Automated} {Reasoning} with {Connection}
  {Calculi} ({AReCCa} 2023)}, volume 3613 of \emph{{CEUR} {Workshop}
  {Proceedings}}, 107--118. Prague, Czech Republic: CEUR.

\bibitem[{Schlichtkrull et~al.(2018)Schlichtkrull, Kipf, Bloem, van~den Berg,
  Titov, and Welling}]{schlichtkrull_modeling_2018}
Schlichtkrull, M.; Kipf, T.~N.; Bloem, P.; van~den Berg, R.; Titov, I.; and
  Welling, M. 2018.
\newblock Modeling {Relational} {Data} with {Graph} {Convolutional} {Networks}.
\newblock In \emph{The {Semantic} {Web} ({ESWC})}, volume 10843 of
  \emph{Lecture {Notes} in {Computer} {Science}}, 593--607. Cham: Springer
  International Publishing.

\bibitem[{Stickel(1988)}]{stickel_prolog_1988}
Stickel, M.~E. 1988.
\newblock A {Prolog} {Technology} {Theorem} {Prover}: {Implementation} by an
  {Extended} {Prolog} {Compiler}.
\newblock \emph{Journal of Automated Reasoning}, 4(4): 353--380.

\bibitem[{Sutcliffe(2017)}]{sutcliffe_tptp_2017}
Sutcliffe, G. 2017.
\newblock The {TPTP} {Problem} {Library} and {Associated} {Infrastructure}.
\newblock \emph{Journal of Automated Reasoning}, 59(4): 483--502.

\bibitem[{Sutton and Barto(2018)}]{sutton_reinforcement_2018}
Sutton, R.~S.; and Barto, A.~G. 2018.
\newblock \emph{Reinforcement {Learning}: {An} {Introduction}}.
\newblock Cambridge, MA: MIT Press, second edition.

\bibitem[{Urban, Vyskočil, and Štěpánek(2011)}]{urban_malecop_2011}
Urban, J.; Vyskočil, J.; and Štěpánek, P. 2011.
\newblock {MaLeCoP} {Machine} {Learning} {Connection} {Prover}.
\newblock In Brünnler, K.; and Metcalfe, G., eds., \emph{Automated {Reasoning}
  with {Analytic} {Tableaux} and {Related} {Methods}}, volume 6793 of
  \emph{Lecture {Notes} in {Computer} {Science}}, 263--277. Berlin, Heidelberg:
  Springer.

\bibitem[{Xu et~al.(2019)Xu, Hu, Leskovec, and Jegelka}]{xu_how_2019}
Xu, K.; Hu, W.; Leskovec, J.; and Jegelka, S. 2019.
\newblock How {Powerful} are {Graph} {Neural} {Networks}?
\newblock In \emph{International {Conference} on {Learning} {Representations}}.

\bibitem[{Zombori, Urban, and Brown(2020)}]{zombori_prolog_2020}
Zombori, Z.; Urban, J.; and Brown, C.~E. 2020.
\newblock Prolog {Technology} {Reinforcement} {Learning} {Prover} ({System}
  {Description}).
\newblock In Peltier, N.; and Sofronie-Stokkermans, V., eds., \emph{Automated
  {Reasoning}}, volume 12167 of \emph{Lecture {Notes} in {Computer} {Science}},
  489--507. Cham: Springer International Publishing.

\end{thebibliography}

\end{document}